\documentclass[lettersize,journal]{IEEEtran}

\usepackage{array}
\usepackage[caption=false,font=normalsize,labelfont=sf,textfont=sf]{subfig}
\usepackage{textcomp}
\usepackage{stfloats}
\usepackage{url}
\usepackage{verbatim}
\AtBeginDocument{\DeclareMathAlphabet{\mathbf}{OT1}{cmr}{bx}{n}}

\usepackage{amsmath,amsfonts}
\usepackage{algorithmic}
\usepackage{multirow,multicol}
\usepackage{color}
\usepackage{graphicx}
\usepackage{graphics}
\usepackage{subfig}
\usepackage{caption}
\usepackage{amsmath}
\usepackage{amsthm}
\usepackage{bbm}
\usepackage{bm}

\makeatletter

\newcommand{\Rmnum}[1]{\expandafter\@slowromancap\romannumeral #1@}
\makeatother

\def\BibTeX{{\rm B\kern-.05em{\sc i\kern-.025em b}\kern-.08em
    T\kern-.1667em\lower.7ex\hbox{E}\kern-.125emX}}
\usepackage{balance}

\begin{document}
\title{A Spatio-temporal Continuous Network for Stochastic 3D Human Motion Prediction }
   
\author{Hua Yu, Yaqing Hou, Xu Gui, Shanshan Feng, Dongsheng Zhou, Qiang Zhang
\thanks{Manuscript created xxx. This work was supported in part by the National Key Research and Development Program of China (No. 2021ZD0112400), the National Natural Science Foundation of China under Grant 62372081, the Young Elite Scientists Sponsorship Program by CAST under Grant 2022QNRC001, the NSFC-Liaoning Province United Foundation under Grant U1908214, the 111 Project, No.D23006, and the Fundamental Research Funds for the Central Universities under grant DUT21TD107, DUT22ZD214. (Corresponding authors: Yaqing Hou and Qiang Zhang.)

Hua Yu, Yaqing Hou, and Qiang Zhang are with the Colleage of Computer Science and Technology, Dalian University of Technology, Dalian 116622, China (e-mail: yhiccd@mail.dlut.edu.cn;	
houyq@dlut.edu.cn; zhangq@dlut.edu.cn)

Shanshan Feng is with the Centre for Frontier AI Research, Institute of High Performance Computing, A*STAR, Singapore. E-mail: victor\_fengss@foxmail.com

Xu Gui and Dongsheng Zhou are with the School of Software Engineering, Dalian University, Dalian 116622, China (e-mail: guixu@s.dlu.edu.cn; zhouds@dlu.edu.cn).
}}

\markboth{Journal of \LaTeX\ Class Files,~Vol.~18, No.~9, September~2020}%
{How to Use the IEEEtran \LaTeX \ Templates}

\maketitle

\begin{abstract}
Stochastic Human Motion Prediction (HMP) has received increasing attention due to its wide applications. Despite the rapid progress in generative fields, existing methods often face challenges in learning continuous temporal dynamics and predicting stochastic motion sequences. They tend to overlook the flexibility inherent in complex human motions and are prone to mode collapse. To alleviate these issues, we propose a novel method called STCN, for stochastic and continuous human motion prediction, which consists of two stages. Specifically, in the first stage, we propose a spatio-temporal continuous network to generate smoother human motion sequences. In addition, the anchor set is innovatively introduced into the stochastic HMP task to prevent mode collapse, which refers to the potential human motion patterns. In the second stage, STCN endeavors to acquire the Gaussian mixture distribution (GMM) of observed motion sequences with the aid of the anchor set. It also focuses on the probability associated with each anchor, and employs the strategy of sampling multiple sequences from each anchor to alleviate intra-class differences in human motions. Experimental results on two widely-used datasets (Human3.6M and HumanEva-I) demonstrate that our model obtains competitive performance on both diversity and accuracy.
\end{abstract}

\begin{IEEEkeywords}
Stochastic human motion prediction, human anchors, mode collapse, smooth human motion sequences.
\end{IEEEkeywords}

\section{Introduction}
\IEEEPARstart{H}{uman} motion prediction (HMP) aims to predict future human motions given a historical human motion sequence, which has achieved significant success across a wide range of vital topics with broad applications, such as virtual reality \cite{yuan2023physdiff,tang2023collaborative}, vision-based human-robot interactions \cite{men2020quadruple,ding2022towards}, and game development \cite{shen2022federated}. 
Previous HMP works focus mainly on the deterministic methods \cite{chen2023hdformer,liu2023multi, yang2022motion, hu2023personalized}, where only one output is produced given an observed human sequence \cite{zheng2022gimo, zhong2022spatio}. Nevertheless, considering the intricate nature of environments and the imperative need for human safety, it is essential for the robot/machine to realize the inherent uncertainty in future human motions.
This underscores the necessity for the designed methods to predict not just a single deterministic future human motion but rather multiple possible future human motions, embracing the concept of stochastic HMP, i.e., a model is designed to fully capture all the multi-pattern motion dynamics from an observed human sequence.
\begin{figure*}
    \centering
    \includegraphics[width=1.0\linewidth]{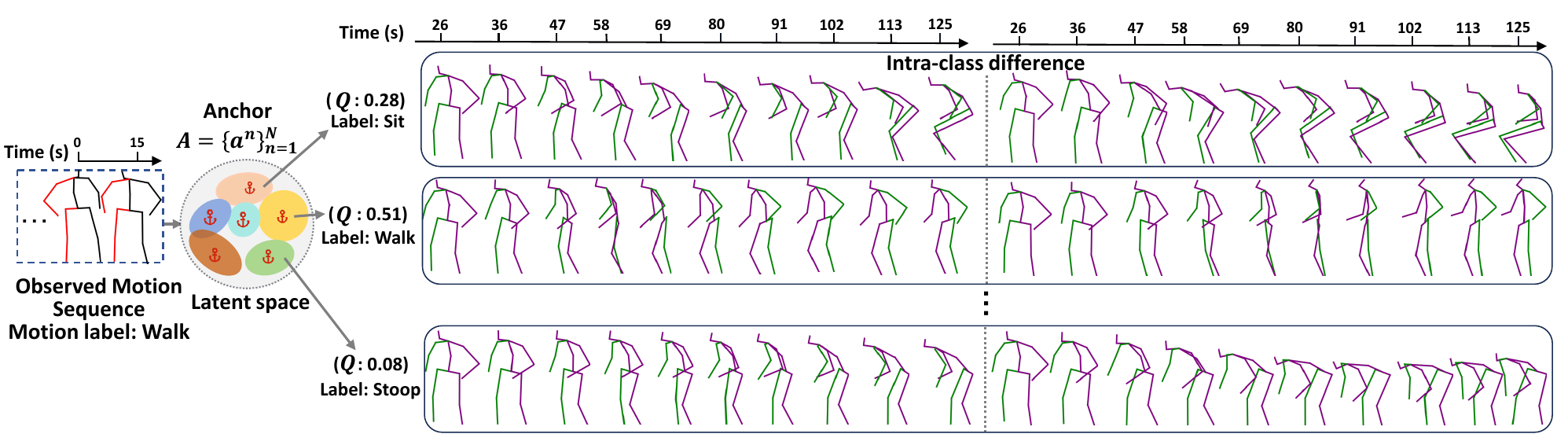}
    \caption{ The example of STCN. Given the observed human motion sequence, STCN aims to predict diverse and smooth human motion sequences. In the latent space, different color denotes different distributions of anchor sets $\bm{A}=\{{a}^{n}\}_{n=1}^{N}$ (human motion patterns). 
    On the right of this figure, different lines denote different human motion patterns. Motions sampled from the same anchor with probabilities $Q$ are depicted by the same line, which can learn the intra-class differences of human motions. }
    \label{fig:FirstImage}
\end{figure*} 

Recent works for stochastic HMP intend to capture multi-pattern motion dynamics with deep generative models, such as variational autoencoders (VAEs) \cite{cai2021unified,mao2021generating,liu2020trajectorycnn,chen2023spatiotemporal}, generative adversarial networks (GANs) \cite{gui2018adversarial,odena2017conditional,ma2021spatial}, and Diffusion Models \cite{barquero2023belfusion,wei2023human}. Specifically, VAEs leverage an encoder network to map the data distribution into a Gaussian latent distribution. This representation of latent space is mapped back to the data distribution using a decoder network.
As depicted in DLow \cite{yuan2020dlow}, they investigated a trainable sampling strategy, enhanced with prior knowledge to promote the diversity of future human motion predictions.
VAEs assume Gaussian distribution for the latent space of the data distribution, which often leads to the generation of blurry human motions. 
GANs-based methods obtain the diversity by randomly sampling a set of latent variables from the latent prior. This usually captures the major pattern with less diversity and ignores the stochastic latent variables, which often suffer from \textit{mode collapse}. 
These methods for stochastic HMP often sample from the densely populated regions of the data distribution while ignoring other motion patterns, resulting in mode collapse.

Another type of generative method is the diffusion-based methods \cite{barquero2023belfusion,song2020score}, which define a diffusion process, in which Gaussian noise is gradually added to the input data using a Markov chain. For example, BeLFusion \cite{barquero2023belfusion} tries to predict human motions via the diffusion model in the latent space. Unfortunately, diffusion-based methods are complex to train and generalize to real-world applications.
Despite their potential performance, existing methods for stochastic HMP utilize discrete-time sequential models, which are not flexible enough to deal with different speeds and frequencies of complex human motions. How to learn \textit{continuous} temporal dynamics remains a challenge.
In addition, these methods overlook intra-class differences when performing the same human motions. Note that each individual exhibits variations when executing the same action, introducing uncertainty to a person performing a certain motion pattern.

To address these challenges, we present a new perspective for stochastic 3D Human Motion Prediction by proposing a novel Spatio-Temporal Continuous Network, namely STCN, to alleviate the mode collapse and enhance the smoothness of predicted diverse motion sequences. STCN is easy to train compared to previous generative methods, which mainly consist of two stages.
In the first stage, to enhance the smoothness of the predicted human motion sequences, we devise a spatio-temporal continuous network based on ordinary differential equation (ODE) to capture the continuous representations of human motions, which utilizes the Quantized Variational Autoencoders (VQVAE) \cite{van2017neural} as the backbone networks.
Furthermore, to alleviate the mode collapse, we innovatively introduce the anchor set to cluster human motion sequences into semantically similar regions. The introduced anchor set contains global human motion information, which functions as priors for distinguishing each human motion pattern. 
In this way, STCN decouples the style of human patterns from the human motion features, alleviating \textit{pattern collapse} and improving the \textit{continuity} of predicted diverse human motion sequences.
In the second stage, the observed motion sequences are encoded into a low-dimensional representation and matched with the learned anchor set to capture potential human motion patterns in the observed sequences.
Then STCN learns the Gaussian Mixture Model (GMM) and the corresponding probabilities of each anchor. Multiple sequences are concurrently sampled from each anchor, which can alleviate \textit{intra-class differences} in human motion.
As shown in Fig.~\ref{fig:FirstImage}, given an observed human motion sequence, STCN can predict stochastic human motion sequences corresponding to potential anchor sets.
In addition, STCN samples multiple sequences from each anchor and predicts the probabilities $Q$, which can mitigate the intra-class differences when performing the same motions.

In summary, the contributions of our work are as follows:
\begin{itemize}
\item{ We propose a novel spatio-temporal ODE-based continuous network, called STCN, to predict stochastic and accurate human motion sequences, which can address discontinuous and mode collapse issues in existing stochastic HMP task.}
\item{ We innovatively introduce the anchor set to denote the human motion patterns, which can facilitate the capture of potential future human motion patterns of observed human motion sequences.}
\item{ We propose to learn the Gaussian mixture model (GMM) distribution and sample multiple sequences of each anchor, which aims to differentiate the intra-class differences between the same motion pattern. }
\item{ The proposed method is comprehensively evaluated on widely-used human motion datasets. The results are analyzed in depth, indicating that the proposed method significantly outperforms the state-of-the-art methods.}
\end{itemize}

The rest of this paper is organized as follows. The related works are briefly introduced in Section II. Then, Section III introduces the proposed method. Next, the experimental design and corresponding analysis are given in Section IV and Section V, respectively. Finally, the conclusion is derived in Section VI.

\section{Related Work}
In this section, we intend to introduce the related work of our method. We first review previous methods for stochastic human motion prediction methods, and analyze the challenges in this task. Then we present existing discrete anchor works, and describe the difference between our method and previous work. 

\subsection{Stochastic Human Motion Prediction.}
Bolstered by the advancements in deep generative methods  \cite{van2017neural,gui2018adversarial}, early efforts on stochastic HMP based on GANs employ generators and discriminators to produce multiple future human motions. \cite{barsoum2018hp} proposed a probabilistic generative adversarial network along with a noise variable to learn a probability density function of future human motions. 
VAEs \cite{mao2021generating,cai2021unified} methods utilize the encoder-decoder mechanism to capture the distribution of the data. These methods usually suffer mode collapse, the major patterns of human motions are more likely to generate samples while ignoring the minor patterns. 
Recent diffusion-based \cite{ho2020denoising,barquero2023belfusion} approaches have been proposed for stochastic HMP. For example, MotionDiffuse \cite{zhang2022motiondiffuse} and MDM \cite{tevet2022human} introduce diffusion models into text-driven motion synthesis. 
MotionGPT \cite{jiang2023motiongpt} builds a unified model and employs a discrete vector quantization for human motion generation.
Diffusion-based methods are usually difficult to train and lack continuous modeling of temporal dynamics, which could only model the discrete-time human motion data.
Neural ODEs can address this issue by parameterizing the derivative of the hidden state using a neural network, which has been applied to many tasks, such as classification \cite{ruiz2023neural} and traffic flow forecasting \cite{fang2021spatial}.
Recent efforts \cite{xing2023hdg,chen2024neural} proposed to utilize the neural ordinary differential equation (NeuralODE) to model the continuous-time relations. However, their method mainly focuses on deterministic HMP and the relations among human joints.
Note that discontinuities are more likely to occur in stochastic HMP which involves diverse future motion sequences. Modeling these processes in discrete time adversely impacts the accuracy of HMP predictions.

To address the mode collapse and model the continuous-time relations of human motions, STCN introduces spatio-temporal ODE-based network to stochastic HMP. Meanwhile pre-training a set of anchors to capture the potential distributions of human motions.

\subsection{Learnable Discrete Anchors.}
Stochastic HMP task aims to capture the full diversity of the potential human motion patterns.
Previous work for stochastic HMP mainly resorts to other generative models \cite{gui2018adversarial,yuan2020dlow,barquero2023belfusion} to enhance the diversity.
For example, Barsoum et al. \cite{barsoum2018hp} proposed a probabilistic generative adversarial network (HP-GAN) along with a noise variable to learn a probability density function of future human motions. However, the method based on GANs often suffer from mode collapse.
Tevet et al. \cite{tevet2022human} designed a Motion Diffusion Model (MDM) for various human motion generation, enabling different modes of conditioning, and different generation tasks. However, their approach is computationally intensive and low-efficiency inference process.

\begin{figure*}[!htb]
    \centering
    \includegraphics[width=6.82
    in]{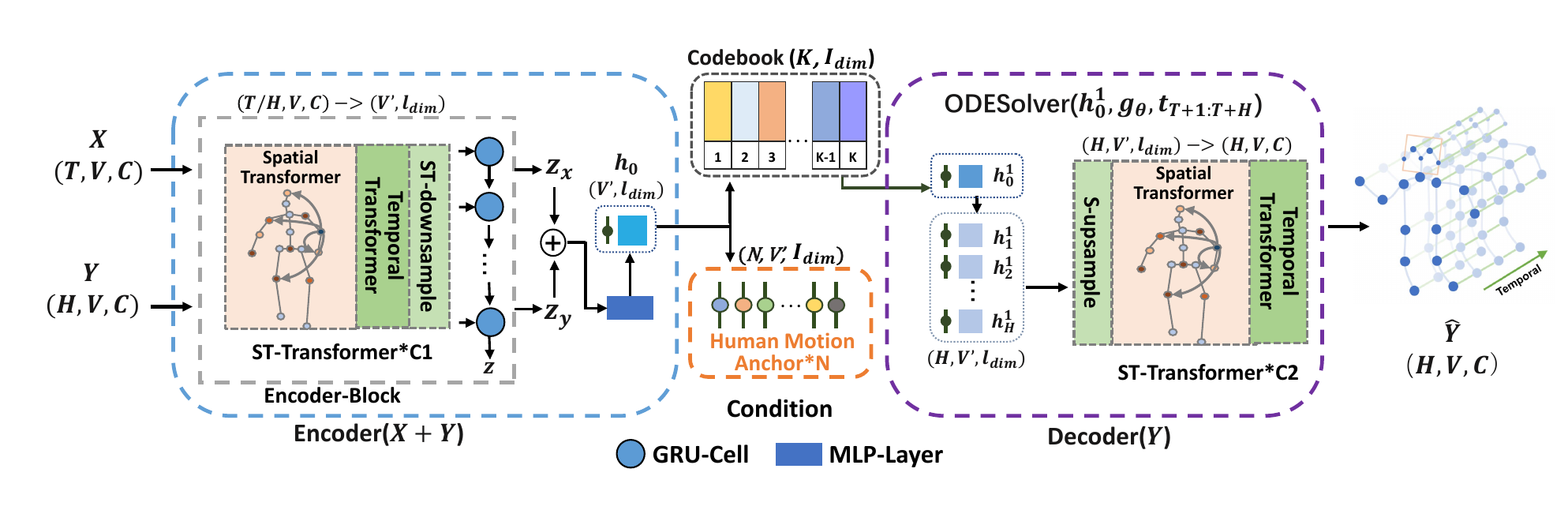} 
    \caption{The first stage of STCN, i.e., human motion reconstruction process. This process aims to capture the anchors (denoted in the orange box) and learn the continuous representations of human motions using an ODE solver in the decoder network (denoted in the purple box).
    }
    \label{fig:vqvae}
\end{figure*}

In this work, we propose to utilize the discrete anchors to capture the diversity of observed human motion sequences and avoid the mode collapse, which is easy to train compared to other generative models. 
The concept of learnable anchors, drawing inspiration from research on utilizing predefined priors and learnable codes in various applications, such as trajectory prediction \cite{chai2019multipath} and object detection \cite{erhan2014scalable}. In these fields, anchors refer to the hypothesis of box candidates that encompass diverse shapes and locations. 

Unlike prior works, our method innovatively incorporates anchors to distinguish different human patterns and prevent mode collapse, allowing the capture of the multi-pattern distribution of human motions. Note that these anchors are not manually crafted, but can be learned via the learning process of human motion continuous representation. 
STCN can learn the probability score and GMM distribution of the corresponding anchors. 

\section{The Proposed Approach}

\subsection{Overview }
\begin{figure*}[!htb]
    \centering
    \includegraphics[width=6.8
    in]{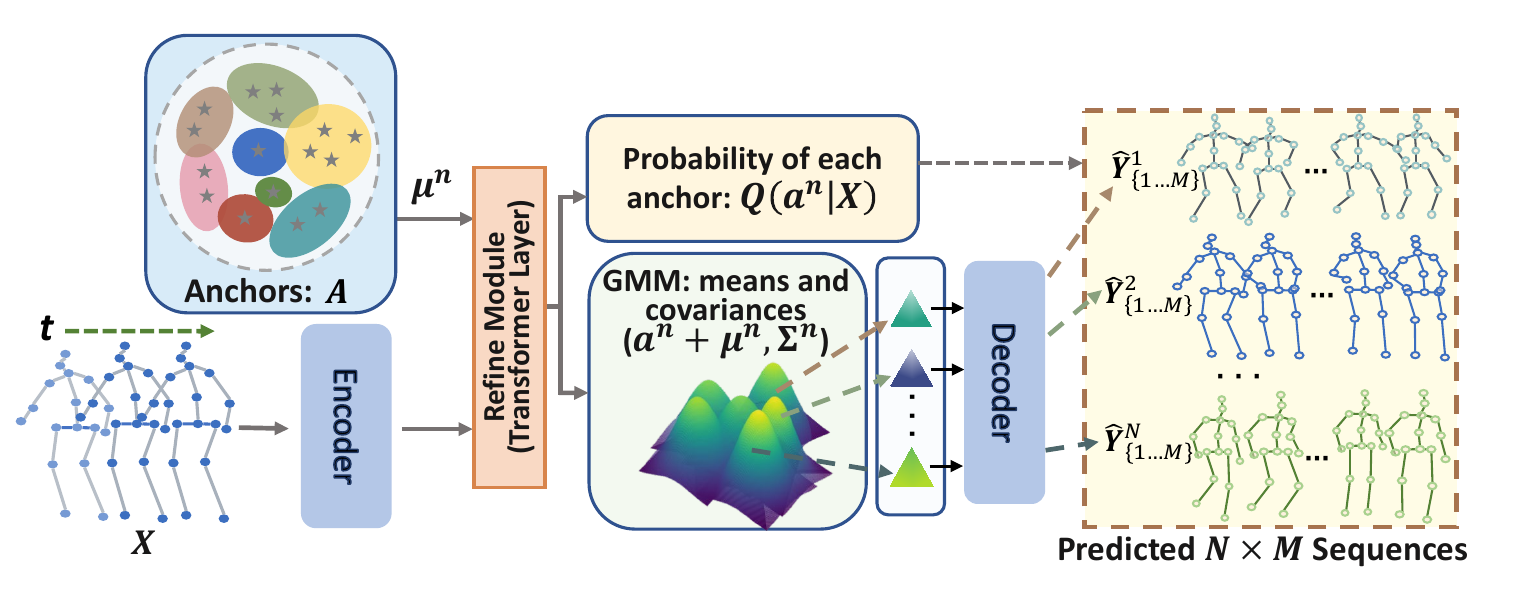} 
    \caption{ The second stage of STCN, i.e., stochastic human motion sequences prediction process. Given an observed human motion sequence $\bm{X}$, we first encode the observed sequence into latent space, then match it with the $\bm{A}=\{{a}^{n}\}_{n=1}^{N}$ to capture potential human patterns. A refine model is utilized to learn the distribution for the observed sequence and the probability $Q$ of each anchor.
    STCN samples from each potential anchor to learn intra-class differences of the same human motions, and employs a decoder to decode them into multiple human motion sequences.}
    \label{fig:APPROACH}
\end{figure*}

Our task is to predict diverse and accurate human motion sequences given the observed human motion sequence. However, there exists \textit{discontinues} and \textit{mode collapse} of the predicted sequences in existing stochastic HMP. 
Therefore, we aim to address these issues through the proposed method STCN, which mainly consists of two stages. 
Fig.~\ref{fig:vqvae} shows the first stage of our method, human motion reconstruction.
In the first stage, STCN aims to learn the anchor sets and continuous motion representations through the human motion reconstruction process (Section B).
Fig.~\ref{fig:APPROACH} shows the second stage of our method, stochastic human motion sequences prediction.
In the second stage, STCN aims to learn the potential human motion patterns of the observed human motion sequences (Section C). In addition, we can sample motion sequences from the learned motion patterns, which can alleviate the intra-class differences.


%

\subsection{Human Motion Reconstruction}
Human motion reconstruction process aims to learn the anchor set corresponding to human motion patterns in a low-dimensional space and continuous flow of human motions. The reconstruction process is based on the Quantized Variational Autoencoders (VQVAE) \cite{van2017neural} network. 
As depicted in Fig.~\ref{fig:vqvae}, the input human motion sequence can be represented by a matrix $\bm{X}+\bm{Y} \in \mathbb{R}^{(T+H) \times D}$, where $T$ and $H$ are the number of frames corresponding to the observed and future motion sequences respectively, $D$ is the dimension of human joint features. 
$\bm{X}+\bm{Y}$ is encoded to low-dimensional latent space features $Z \in \mathbb{R}^{V^{\prime}\times l_{dim}}$ by encoder network $ {\rm Enc}(\cdot)$, which is implemented by the Transformer and GRU module.
Each row of $Z$, represented by $z_i$, is substituted by its closest vector $k_j$ using a quantization codebook: 
\begin{equation}
\hat{z_i}=\arg\min_{k_j\in {K}}||z_i-k_j||.
\end{equation}
Afterwards, the quantized feature $\hat{z}_i$ is decoded to $\hat{\bm{Y}}$ by the decoder network ${\rm Dec}(\cdot)$.

Note that traditional neural networks for human motion prediction typically operate on discrete time steps, which are not flexible enough to deal with different speeds and frequencies of complex human motions.
ODE learns a continuous dynamical function, ensuring that the predicted motion sequences are smooth over time, which is crucial for the naturalness and continuity of human motions. Human motion can be viewed as a dynamic system.
Therefore, in the decoder network, a spatio-temporal ODE-based network is designed to model continuous dynamics of human motions, which utilizes neural network $f_\theta$ with trainable parameters $\theta$ to model the derivative of the latent state $h^1_t$ at a given time $t$. The latent space $Z$ is transformed to the initial value $h^1_0$ of ODESolver. This process is denoted as follows:
\begin{equation}
    \begin{aligned}
    \frac{dh^1_t}{dt}=f_\theta(h^1_t,t), 
    h^1_t=h^1_0 + \int_0^tf_\theta(h^1_t,t)dt,
    \end{aligned}
\end{equation}
where $h^1_0$ is the initial value. To distinguish from the initial values, the first stage is labeled as $h_0^1$, and the second stage is $h_0^2$. ODESolver is the numerical ODESolver given equation $\frac{dh^1_t}{dt}=g_{\tau}$, which is implemented by Transformer. $f_\theta$ is implemented with spatio-temporal transformer ODE-based network to facilitate the continuous reconstruction of human motions. 
Formally, the formulation of the decoder network is formulated as:
\begin{equation}
\begin{aligned}
h^1_{1:H}&=\mathrm{ODESolver}(g_\tau, h^1_0, t^1_{1:H}),\\
\hat{\bm{Y}} &= {\rm Dec}( h^1_{1:H} \mid Z_x ).
\end{aligned}
\end{equation}

The decoder aims to generate the target human motion sequences based on the latent $h^1_{1:H}$ output by the decoder. 
$Z_x$ is utilized as conditions in the decoder to enhance the continuity between the observed and predicted sequences, and address the issue of disappearing gradients.
The spatio-temporal ODE-based network is learned by minimizing the following loss function:
\begin{equation}
\label{equ:vq}
\begin{aligned}
\mathcal{L}_{VQ}=& \mathcal{L}(\bm{Y},\hat{\bm{Y}}) 
+||\hat{Z}-\operatorname{sg}({Z})||_{2}^{2}+\beta||\operatorname{sg}(\hat{Z})-Z||_{2}^{2},
\end{aligned}
\end{equation}
where $\mathrm{sg}[\cdot]$ is the stop gradient operator to prevent the gradient from backpropagating through its operand. The first term $\mathcal{L}(\bm{Y},\hat{\bm{Y}}) = \sum_{t=1}^{H}\sum_{v=1}^{V}\parallel \bm{y}^{(v)}_t-\hat{\bm{y}}^{(v)}_t\parallel_2$ represents the reconstruction error. The second term aims to optimize the codebook by pushing the $\hat{Z}$ and its associated ${Z}$ close to the output of the encoder $Z$. The last term intends to optimize the encoder by pushing $Z$ close to its nearest latent vector in the codebook.

The above process describes the learning procedure of continuous human motion representation. 
Previous methods for stochastic HMP tend to suffer from mode collapse, in this work, we propose a simple yet efficient way to alleviate this issue.
Specifically, the encoder network encodes the motion sequences into latent space, ${\rm Enc}(\bm{X+Y}) \rightarrow Z$. 

\subsection{Human Anchor Loss}
In this paper, the proposed method aims to capture the full diversity of the potential human motion patterns using the introduced human anchors. 
Therefore, in the first stage of the proposed method, after capturing the low-dimensional latent space features, we employ the K-means algorithm to split the latent space into $N$ clusters, thereby we can obtain $N$ anchors $\boldsymbol{A}=\{{a}^{n}\}_{n=1}^{N}$. The learned anchor set denotes the human motion patterns, which are utilized to capture the potential motion patterns corresponding to the observed motion sequence in the second stage. This operation aims to avoid the mode collapse issue.

Based on the above process, we propose the human anchor loss function to predict the potential human motion patterns more accurately. The formula is as follows: 
\begin{equation}
L_{anchor}= \min \mathcal{D}^2[{\rm FC}(a^n+\mu^n)-h_0^1],
\end{equation}
where $\mathcal{D}(\cdot,\cdot)$ is the Euclidean distance, ${\rm FC}(a^n+\mu^n)$ denotes the transformed mean of each anchor through fully connected (FC) layer,
$h_0^1$ represents the mean of $\bm{Y}$ sequence at the initial time of the first stage.

\subsection{Stochastic Human Motion Sequences Prediction}
In the first stage, we have learned the continuous representation and the anchor set of human motion sequences. In the second stage, we aim to utilize this information to learn the potential human motion patterns of the observed motion sequence, thereby enhancing the diversity and continuity of future human motion sequences.

Specifically, given an observed human motion sequence $\bm{X} = \left\{\textbf{x}_1, \textbf{x}_2, \ldots, \textbf{x}_T \right\}$ with the length of $T$, where $\textbf{x}_t \in \mathbb{R}^{V \times C}$ is denoted by 3D coordinates at time $t$, $C$ denotes the 3D coordinates of human joints ($C=3$), $V$ is the number of human joints in a frame.
STCN utilizes an encoder to embed the observed sequence into latent space, and matches them with the learned number of $N$ Anchors, $\bm{A} = \{a^n\}^N_{n=1}$ to capture the potential human motion patterns. The encoder and decoder network used in the second stage have the same architecture as the first stage. 
A refine module implemented by the transformer is introduced to learn the distribution $\phi$ of the observed sequence. Meanwhile, STCN outputs the corresponding probabilities $Q$ of each anchor, i.e., the probabilities of possible future motion patterns corresponding to a certain motion pattern given the observed motion sequence, which are formulated as follows:
\begin{equation}
    Q({a}^{n}|\boldsymbol{X})=\frac{\exp r_{n}(\boldsymbol{X})}{\sum_{i=1}^N\exp r_{i}(\boldsymbol{X})}
\end{equation}
where $r_i(\boldsymbol{X}): \mathbb{R}^{d(\boldsymbol{X})} \mapsto \mathbb{R}$ is the output of a deep neural network.
The GMM distribution $\phi$ of the latent space is formulated as:
    \begin{equation}
      \phi(Z|a^n,\boldsymbol{X})=\mathcal{N}(Z|a^n+\mu^n(\boldsymbol{X}),\Sigma^n(\boldsymbol{X})),
    \end{equation}
where $a^n+\mu^n(\boldsymbol{X})$ is the mean of the distribution, which represents the offset $\mu^n$ from the anchor $a^n$. 
$\Sigma^n(\boldsymbol{X})$ denotes the distribution covariance. 
The distribution across the entire motion sequence can be formulated as follows:
    \begin{equation} 
        p(Z|\boldsymbol{X})=\sum_{n=1}^N Q(a^n|\boldsymbol{X})\phi(Z|a^n,\boldsymbol{X}).
    \end{equation}

Finally, a decoder network is leveraged to decode these latent space variables to diverse human motion sequences. Our method outputs a total of $S$ human action sequences, which denoted as $\hat{\bm{Y}}_{1},\hat{\bm{Y}}_{2},\ldots,\hat{\bm{Y}}_{S}$, $ S =N \times M$, $N$ denotes the number of anchors, and $M$ is the number of samples deviating from the same anchor.

\begin{table*}[ht!]
\caption{The comparison results between the proposed method and state-of-the-art methods on Human3.6M and HumanEva-I datasets. The best results are in bold. Lower is better for all metrics except the APD metric. } 
\label{table1}
\resizebox{\linewidth}{!}{ 
\begin{tabular}{c|c|ccccc|ccccc} 
\hline
\multirow{2}{*}{Type}          & \multirow{2}{*}{Method} & \multicolumn{5}{c|}{Human3.6M}                                                      & \multicolumn{5}{c}{HumanEva-I}                                                      \\ \cline{3-12} 
     &     & APD $\uparrow$    & ADE $\downarrow$  & FDE $\downarrow$    & MMADE $\downarrow$         & MMFDE $\downarrow$    
           & APD $\uparrow$    & ADE $\downarrow$  & FDE $\downarrow$    & MMADE $\downarrow$         & MMFDE $\downarrow$         \\ \hline
\multirow{3}{*}{Deterministic} & LTD(ICCV'19)     & 0.000    & 0.516     & 0.756          & 0.627          & 0.795          & 0.000           & 0.415          & 0.555          & 0.509          & 0.613          \\
                               & MSR(ICCV'21)            & 0.000         & 0.508 & 0.742          & 0.621          & 0.791          & 0.000           & 0.371          & 0.493          & 0.472          & 0.548          \\ 
                               & acLSTM (ICLR’18)      & 0.000           & 0.789    &1.126          & 0.849          & 1.139             & 0.000         & 0.429          & 0.541             & 0.530          & 0.608          \\
                               \hline
\multirow{12}{*}{Stochastic}    & Pose-Konws(ICCV'17)     & 6.723           & 0.461          & 0.560          & 0.522          & 0.569          & 2.308           & 0.269          & 0.296          & 0.384          & 0.375          \\
              & MT-VAE(ECCV'18)         & 0.403           & 0.457          & 0.595          & 0.716          & 0.883          & 0.021           & 0.345          & 0.403          & 0.518          & 0.577          \\
             & GMVAE(arKiv'16)         & 6.769           & 0.461          & 0.555          & 0.524          & 0.566          & 2.443           & 0.305          & 0.345          & 0.408          & 0.410          \\ \cline{2-12} 
             & DLow(ECCV'20)           & 11.741          & 0.425          & 0.518          & 0.495          & 0.531          & 4.855           & 0.251          & 0.268          & 0.362          & 0.339          \\
             & HP-GAN(CVPRW'18)        & 7.214           & 0.858          & 0.867          & 0.847          & 0.858          & 1.139           & 0.772          & 0.749          & 0.776          & 0.769          \\
             & GSF(ICLR'19)            & 9.330           & 0.493          & 0.592          & 0.550          & 0.599          & 4.538           & 0.273          & 0.290          & 0.364          & 0.340          \\
             & GSPS(ICCV'21)           & 14.757 &0.389 &0.496 &0.476 &0.525 &5.825 &0.233 &0.244 &0.343 &0.331          \\
              & MOJO(CVPR'21)           & 12.579          & 0.412          & 0.514          & 0.497          & 0.538          & 4.181           & 0.234          & 0.244          & 0.369          & 0.347          \\
              & DivSamp(ACM MM'22)      & 15.310          & 0.370          & 0.485          & 0.475          & 0.516          & 24.724 & 0.564          & 0.647          & 0.623          & 0.667          \\
             & BeLFusion(CVPR'23)     & 7.602           & 0.372          & 0.474 & 0.473          & 0.507          & 9.376           & 0.513          & 0.560          & 0.569          & 0.585          \\ 
             & MotionDiff(AAAI'23)     & 15.353           & 0.411          & 0.509 & 0.508          & 0.536          & 5.931           & 0.232          & 0.236          & 0.352          & 0.320          \\ \cline{2-12} 
             & \textbf{STCN(Ours)}        & \textbf{17.512} & \textbf{0.242} & \textbf{0.471}          & \textbf{0.301} & \textbf{0.317} & \textbf{26.894 }         & \textbf{0.103} & \textbf{0.106} & \textbf{0.213} & \textbf{0.161} \\ \hline
\end{tabular} }
\end{table*}

In addition, we utilize the following loss function to optimize the model parameters to get the specific parameters of the distribution:

\textbf{Negative Log-likelihood (NLL) Loss.} 
The proposed method is trained by maximizing the log-likelihood of the human motions. The distribution parameters $Q(\mathbf{a}^{n}|\boldsymbol{X})$, $a^n+\mu^n(\bm{X})$, $\Sigma^n(\bm{X})$ are parameterized by weights $\varphi$ with the following negative log-likelihood loss:
\begin{equation} 
\begin{aligned}
    L_{nll}(\varphi)=&-\sum_{m=1}^{M}\sum_{n=1}^{N}\mathbbm{1}(k=\hat{k}^m)\Big[logQ(\mathbf{a}^{n}|\bm{X^m};\varphi)+ \\
    &\log\mathcal{N}(s^n|a^{n}+\mu^{n},\Sigma^{n};\bm{X}^m;\varphi)\Big],
\end{aligned}
\end{equation}
where $\hat{k}^m$ is the index of the anchor most closely matching the ground truth human motion sequence, and $\mathbbm{1}$ is the indicator function. 

\textbf{Reconstruction Loss.} 
To ensure the accuracy of the predicted future human motions, the reconstruction loss $L_{re}$ is introduced into our work.
The dataset contains only one ground-truth future motion for each observed sequence, it is noted that several motion sequences have similar past motions. For each observed sequence, a distance threshold is set to search for the training samples with similar past motions, then their future motion is treated as the pseudo ground truth.
Supposing that $\{Y_p\}^P_{p=1}$ represents the pseudo-label, the reconstruction loss is defined as follows: 
\begin{equation}
    L_{re}=\frac{1}{P}\sum_{p=1}^{P}\min_{j}\|\hat{\bm{Y}}_{j}-\bm{Y}_{p}\|^{2},
\end{equation}
where $j\in \{1,2,\cdots, S\}$. 

In summary, the loss function of the second stage is the weighted sum of the above-mentioned loss functions. The formula of the loss function $L_{total}$ is expressed as follows:
\begin{equation}
L_{total}=\alpha_1L_{nll}(\varphi)+\alpha_2L_{anchor}+\alpha_3L_{re},
\end{equation}
where the parameters $\alpha_1$, $\alpha_2$, and $\alpha_3$ are the coefficients associated with different losses, which are set empirically in the experiments. Specifically, the loss coefficients are set to $\alpha_1=0.4$, $\alpha_2=0.3$ and $\alpha_3=0.3$, respectively.


\section{Experimental Design }
In this section, we mainly describe the experimental design, including the used human motion datasets, parameter settings, evaluation metrics, and baseline methods.
\subsection{Datasets}
The experiments are conducted on two widely used motion capture datasets, i.e., Human3.6M \cite{ionescu2013human3} and HumanEva-I \cite{sigal2010humaneva}. The details of the two datasets are described as follows.

\textbf{Human3.6M} is a widely utilized dataset for 3D human motion prediction, featuring 15 motions performed by 7 actors. We preprocess the dataset through a down-sampling operation to enhance the training efficiency, reducing the frame rate (in frames per second, fps) from 50 to 25. This down-sampling operation aims to make the inherent motion patterns clearer.
We predict 100 future frames given 25 observed frames.
The proposed method is trained on five subjects (S1, S5, S6, S7, and S8) and is then tested on two subjects (S9 and S11).

\textbf{HumanEva-I} contains 3 subjects recorded at 60 Hz, each of which performs 5 action categories. The pose is represented by 15 joints. The proposed method predicts 60 future poses (1s, 60fps) given 15 past frames.
\subsection{Parameter Settings}
For the embedding of the observed human motion sequence, the batch size is 128, and the number of used human joints is set to 16. For VQVAE, the codebook size is set to 512$\times$512. The number of anchors ($N$) is set to 20. The number of samples ($M$) is set to 5. The proposed method is implemented based on the PyTorch framework in Python 3.6. 
In the first stage, the input is $\bm{X}+\bm{Y} $. For Human3.6M, the total frame of input sequence t-total is set to 125, and for HumanEva-I, t-total is set to 75.
In the second stage, the input is $\bm{X}$.
t-{pred} denotes the number of poses to be
predicted. For the Human3.6M dataset, t-{pred} is set to 100, and for the HumanEva-I dataset, t-{pred} is set to 60.
To guarantee the convergence of the proposed method, the Adam optimizer is adopted to train our model. The learning rate is initially set to $10^{-4}$ with a 0.98 decay every 10 epochs. The proposed method is trained for 500 epochs for both training and evaluation. 
All the training and testing processes are conducted on NVIDIA A100.

\subsection{Metrics}
In the experiment, for a fair comparison, our method employs the following evaluation metrics, which are consistent with those suggested by  \cite{yuan2020dlow}: APD, ADE, FDE, Multi-Modal ADE (MMADE), and Multi-Modal FDE (MMFDE) metrics. The details are as follows:
\begin{itemize}	
\item[$1)$] APD: Average Pairwise Distance between all pairs of motion samples defined as $\frac{1}{N(N-1)} \sum_{i=1}^N \sum_{j \neq i}^N\left\|\mathbf{Y}_i-\mathbf{Y}_j\right\|_2$. 
\item[$2)$] ADE: Average Displacement Error over the whole sequence between the ground truth and the closest generated motion defined as $\frac{1}{f} min_i \left\|\mathbf{Y}_i-\mathbf{X}\right\|_2 $.
\item[$3)$] FDE: Final Displacement Error between the last frame of the ground truth and the closest motion’s last frame defined as $ min_i \left\|\mathbf{Y}_i[f] -\mathbf{X}[f]\right\|_2 $.
\item[$4)$] Multi-Modal ADE and FDE metrics (MMADE and MMFDE) are the multimodal versions of ADE and FDE: ``Similar'' motions are grouped by using an arbitrary distance threshold at t = 0 and the average metric over all these grouped motions is reported. Similar to FDE, MMFDE  only calculates the error of end poses. 
\end{itemize}	
Note that APD is utilized to measure the diversity while others are utilized to measure the accuracy of the proposed method.
For ADE and FDE, we sample 50 motions in the distribution that correspond to the highest probability score.
For MMADE and MMFDE, we select the top five distributions according to probabilities, and each distribution samples 50 motions. 
\subsection{Baseline Methods}
To evaluate the accuracy and diversity of the proposed method, we compare STCN with two categories of state-of-the-art approaches.
(1) Deterministic motion prediction methods: including LTD \cite{mao2019learning}, MSR \cite{dang2021msr} and acLSTM \cite{li2018structure}. These methods mainly focus on predicting one motion sequence. We compare the accuracy performance of predicted sequences for STCN with these methods in the experiment.
(2) Stochastic motion prediction methods, including Pose-Konws \cite{walker2017pose}, MT-VAE \cite{yan2018mt}, GMVAE \cite{dilokthanakul2016deep}, DLow \cite{yuan2020dlow}, HP-GAN \cite{barsoum2018hp}, DSF \cite{yuan2019diverse}, MOJO \cite{zhang2021we}, DivSamp \cite{dang2022diverse}, BeLFusion \cite{barquero2023belfusion}, and MotionDiff \cite{wei2023human}. These stochastic methods include VAE, GAN, and diffusion-based methods. We compare the diversity and accuracy of STCN with these methods in the experiment. 

\begin{table}[!htp]
\centering
\caption{Ablation studies results on Human3.6M dataset in terms of different anchors of STCN. }
\label{ablation1}
\resizebox{\linewidth}{!}{
\begin{tabular}{c|cccccc}
\hline
Anchor & APD $\uparrow$    & ADE $\downarrow$  & FDE $\downarrow$    & MMADE $\downarrow$         & MMFDE $\downarrow$ \\ \hline
$N$=0    & 0.000     &  0.702   & 0.753    & 0.658   & 0.741        \\ 
$N$=5    & 10.842     & 0.590    & 0.595  & 0.614   &  0.682        \\ 
$N$=10    & 12.901     & 0.461    & 0.681   & 0.501   &  0.549       \\ 
$N$=15    & 14.032     & 0.449    & 0.605   & 0.498   &  0.526        \\ 
\textbf{$N$=20}   & \textbf{17.512} & \textbf{0.242} & \textbf{0.503}          & \textbf{0.301} & \textbf{0.317}    \\ 
$N$=30    & 16.491     & 0.332    & 0.501    & 0.403    & 0.401      \\ 
$N$=40    &16.502 &0.342 &0.513 &0.412 &0.402      \\
\hline
\end{tabular} }
\end{table}

\begin{table}[!htp]
\centering
\caption{Ablation studies results on HumanEva-I dataset in terms of the influence of different anchors of the proposed method. }
\label{humaneva}
\resizebox{\linewidth}{!}{
\begin{tabular}{c|cccccc}
\hline
Anchor & APD $\uparrow$    & ADE $\downarrow$  & FDE $\downarrow$    & MMADE $\downarrow$         & MMFDE $\downarrow$ \\ \hline
$N$=0    & 0.000     &  0.682   & 0.563    & 0.549   & 0.740        \\ 
$N$=5    & 19.813    & 0.410    & 0.505  & 0.504   &  0.562        \\ 
$N$=10    & 23.041     & 0.209    & 0.312   & 0.411   &  0.549       \\ 
$N$=15    & 24.492     & 0.149    & 0.205   & 0.308   &  0.526        \\ 
\textbf{$N$=20 }  & \textbf{26.894} & \textbf{0.103} & \textbf{0.106}          & \textbf{0.213} & \textbf{0.161}    \\ 
$N$=30    & 26.031     & 0.107    & 0.114    & 0.232    & 0.191         \\ 
$N$=40   & 25.921     & 0.202    & 0.209    & 0.203    & 0.164         \\ \hline
\end{tabular} }
\end{table}

\begin{table}[htpb]
\centering
\caption{Ablation studies results on Human3.6M dataset in terms of the NeuralODE, anchor set, and different modeling methods. }
\label{ablation2}
\resizebox{\linewidth}{!}{
\begin{tabular}{c|cccccc}
\hline
Method & APD $\uparrow$    & ADE $\downarrow$  & FDE $\downarrow$    & MMADE $\downarrow$         & MMFDE $\downarrow$ \\ \hline
w/o NeuralODE   & 14.013    & 0.312   & 0.596    &  0.613   & 0.402     \\ 
w   NeuralODE   & \textbf{17.512} & \textbf{0.242} & \textbf{0.503}          & \textbf{0.301} & \textbf{0.317}             \\         \hline
w/o anchor set   & 14.604    & 0.406   & 0.616    &  0.703   & 0.411     \\ 
w   anchor set   & \textbf{17.512} & \textbf{0.242} & \textbf{0.503}          & \textbf{0.301} & \textbf{0.317}  \\ \hline
Bingham     & 15.885    & 0.396   & 0.597    &  0.524   & 0.415  \\ 
Laplace     & 15.981    & 0.391   & 0.596    &  0.521   & 0.411 \\ 
GMM          & \textbf{17.512} & \textbf{0.242} & \textbf{0.503}          & \textbf{0.301} & \textbf{0.317}  \\ \hline
\end{tabular}  }
\end{table}

\begin{table*}[htpb]
\centering
\caption{ Comparison results on HumanEva-I considering the influence of NeuralODE. }
\label{NeuralODE}
\begin{tabular}{ccccccc|cc}
\hline
\multirow{2}{*}{Type}          & \multirow{2}{*}{Solvers} & \multirow{2}{*}{Order} & \multicolumn{4}{c|}{Short-term}                               & \multicolumn{2}{c}{Long-term} \\
                               &                          &                        & 80            & 160           & 320           & 400           & 560           & 1000          \\ \hline
Fixed-step                     & Euler                    & 1                      & 0.30          & 0.58          & 0.65          & 0.91          & 1.04          & 1.25          \\
                               & RK4                      & 4                      & 0.29          & 0.49          & 0.57          & \textbf{0.85} & \textbf{1.02} & 1.26          \\
                               & Explicit Adams           & 4                      & 0.27          & 0.48          & 0.58          & 0.97          & 1.14          & 1.30          \\
                               & Implicit Adams           & 4                      & 0.28          & 0.51          & 0.63          & 1.13          & 1.06          & 1.32          \\ \hline
\multirow{2}{*}{Adaptive-step} & Fehlberg2                & 2                      & 0.31          & 0.49          & 0.61          & 1.09          & 1.05          & 1.32          \\
&bosh3   &3   &0.26    &0.48    &0.57    &0.94    &1.03    &1.25   \\
                               & DOPRI5                   & 4                      & \textbf{0.25} & \textbf{0.47} & \textbf{0.56} & 0.86          & 1.07          & \textbf{1.24} \\ \hline
\end{tabular}
\end{table*}
\section{Results and Analysis}

\subsection{Quantitative Results}
Table \ref{table1} summarizes the comparison results of the proposed method and baseline methods on the Human3.6M and HumanEva-I datasets. From the empirical evidence, it is observed that the proposed method consistently outperforms all the baselines based on all the evaluation metrics. 
Specifically, for the deterministic approaches, the diversity metric results are 0 since these approaches only predict one sequence. The prediction accuracy of these approaches is inferior to the stochastic approaches. It is speculated that the deterministic prediction approaches tend to predict an average pattern, causing higher errors.
For the stochastic approaches, STCN outperforms the state-of-the-art methods by a large margin in terms of diversity and accuracy metrics. This is mainly because STCN can better capture the full diversity of the data than the baseline. These results indicate that the introduced predefined anchors facilitate the differentiation of human motion patterns while diversity is increasing. Moreover, the sampled $M$ human motion sequences from the GMM of each observed sequence can not only facilitate the accuracy of the predicted human motions but also alleviate intra-class differences when performing the same human motion. 

\begin{figure}[!htb]
    \centering
    \includegraphics[width=1.0\linewidth]{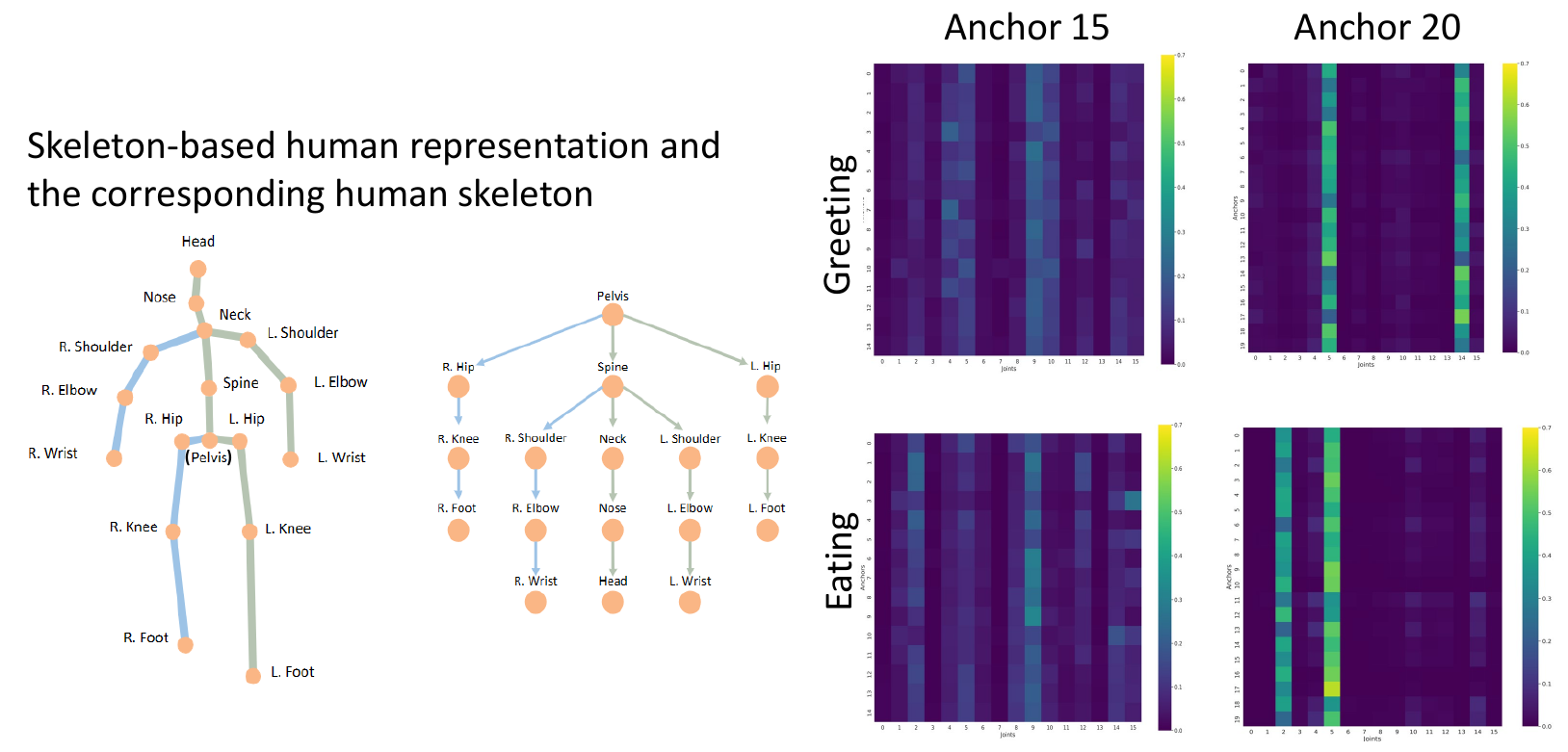}
    \caption{  Visualization of the attention maps of the different anchors. The left side of this figure is the annotation of used human body joints in the HumanEva-l dataset. The right side of the figure visualizes the changes of attention maps under different anchors. } 
    \label{fig:attention}
\end{figure}

\begin{figure}[htp]
    \centering    
    \includegraphics[width=0.75\linewidth]{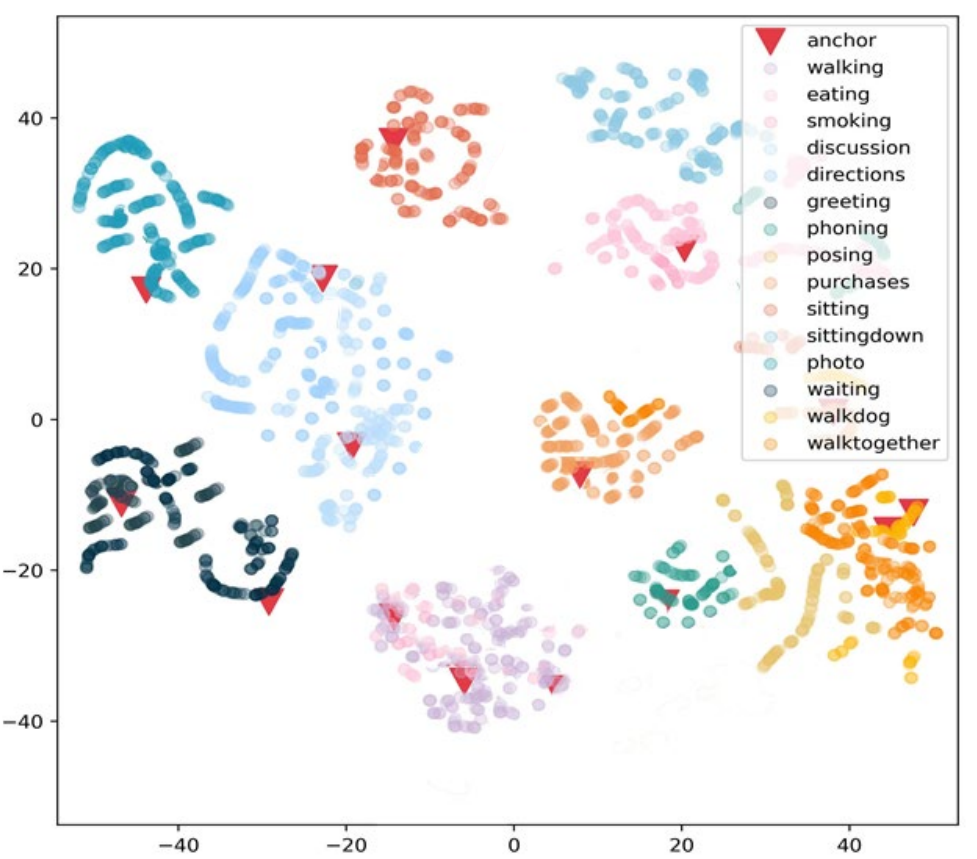}
    \caption{ t-SNE plot of the visualization of the potential human motion patterns. Note that some motions, such as ``Discussion'' and ``Directions'', are very hard to identify and are thus spread over other motions. Other motions, such as ``Walking'' and ``Walking with dog'', overlap due to their similarity. }
    \label{fig:t-SNE}
\end{figure}

\begin{figure*}[!htpb]
    \centering
    \includegraphics[width=5.01in]{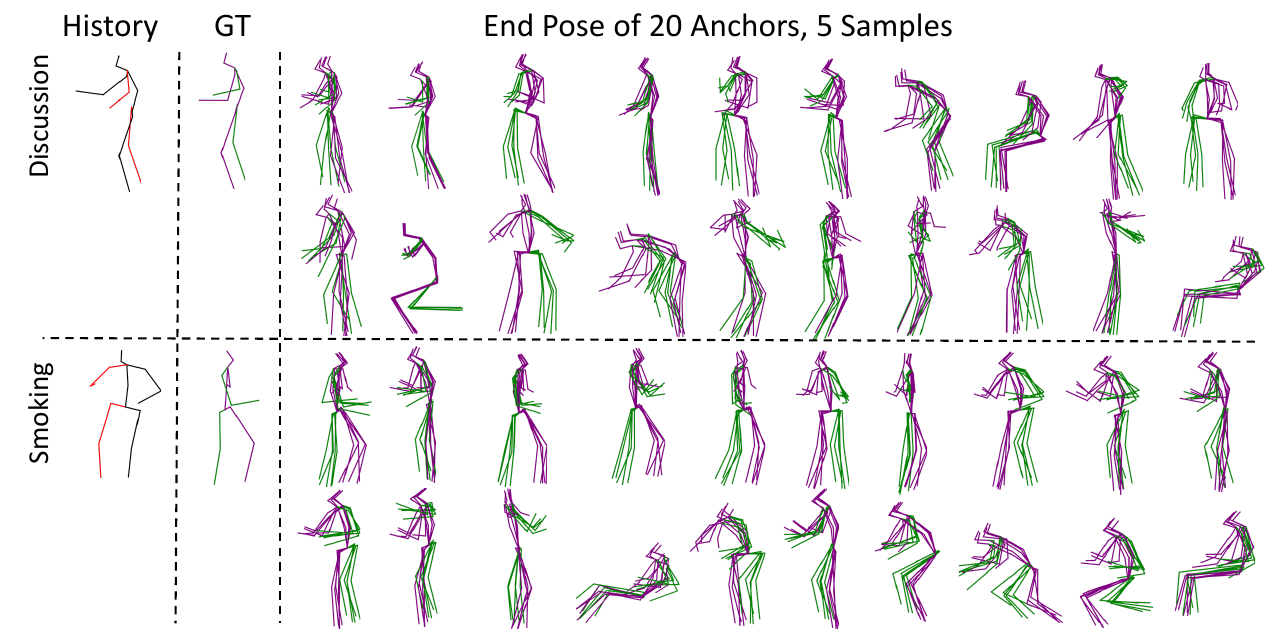} 
    \caption{
    Visualization of the ``Discussion'' and ``Smoking'' human motions from the Human3.6M dataset, which shows the end poses of 20 anchors for human motion, with the number of samples set to 5. These sequences show the captured potential human motion patterns and intra-class differences between human motions. 
    }
    \label{fig:Sample}
\end{figure*}

\subsection{Ablation Studies}
Ablation studies are conducted to justify the contributions of each module in the proposed method STCN.
In the ablation studies, we first evaluate the influence of different numbers of anchors on the results. The comparison results on Human3.6M are reported in Table \ref{ablation1}. It can be observed that there is an intrinsic trade-off between diversity and accuracy. Setting the anchor to 0 indicates that the proposed method is a deterministic approach, and the APD is 0.
More anchors indicate that the proposed model has a larger chance of covering multiple human patterns. However, when diversity exceeds a certain level, the trade-off between diversity and accuracy becomes noticeable. Finally, the number of anchors is set to 20 to enable the model to be more expressive.

In addition, we also conduct ablation studies to validate the influence of different numbers of anchors $N$ on the HumanEva-I dataset. The comparison results are reported in Table \ref{humaneva}. 
Extensive experiments have shown that during the anchor learning process, when the number of anchors is too low, it proves insufficient for covering potential human motion patterns of the observed human motion sequence. Conversely, an excessive number of anchors can lead to the emergence of repetitive motion patterns. Anchors are learned regarding different datasets. Only when the optimal number of anchors is learned, the designed model can achieve a balance between diversity and accuracy.

Moreover, the experiments analyzed the impact of the introduced NeuralODE, anchor set, and different modeling methods of STCN. The comparison results are shown in Table \ref{ablation2}.
Firstly, we noticed that introducing NeuralODE into this task has a significant improvement in the accuracy metrics compared to traditional discrete-time models without (w/o) NeuralODE module. 
NeuralODE can provide predictions at any time for stochastic future motion sequences due to the property of ODE.
For the introduced anchor set, we noticed that STCN exhibits a decrease in diversity metrics when implemented without anchor set, leading to a reduction in the model's generalization capability. These results validate that the introduced anchor set can alleviate the mode collapse in an efficient way.
Moreover, we also evaluate the influence of using different modeling methods of the human motion sequences. We replace the Gaussian mixture model (GMM) distribution with Laplace and Bingham distributions. Unlike the joint position representation, Laplace and Bingham distributions are designed to handle rotations and more complicated quaternion representations. Using GMM distribution for human motions reports a superior performance under these metrics.
The comparison results indicate that the Gaussian mixture model distribution is more favorable to human motion prediction.

\begin{figure*}[htbp]
    \centering
    \includegraphics[width=6.0in]{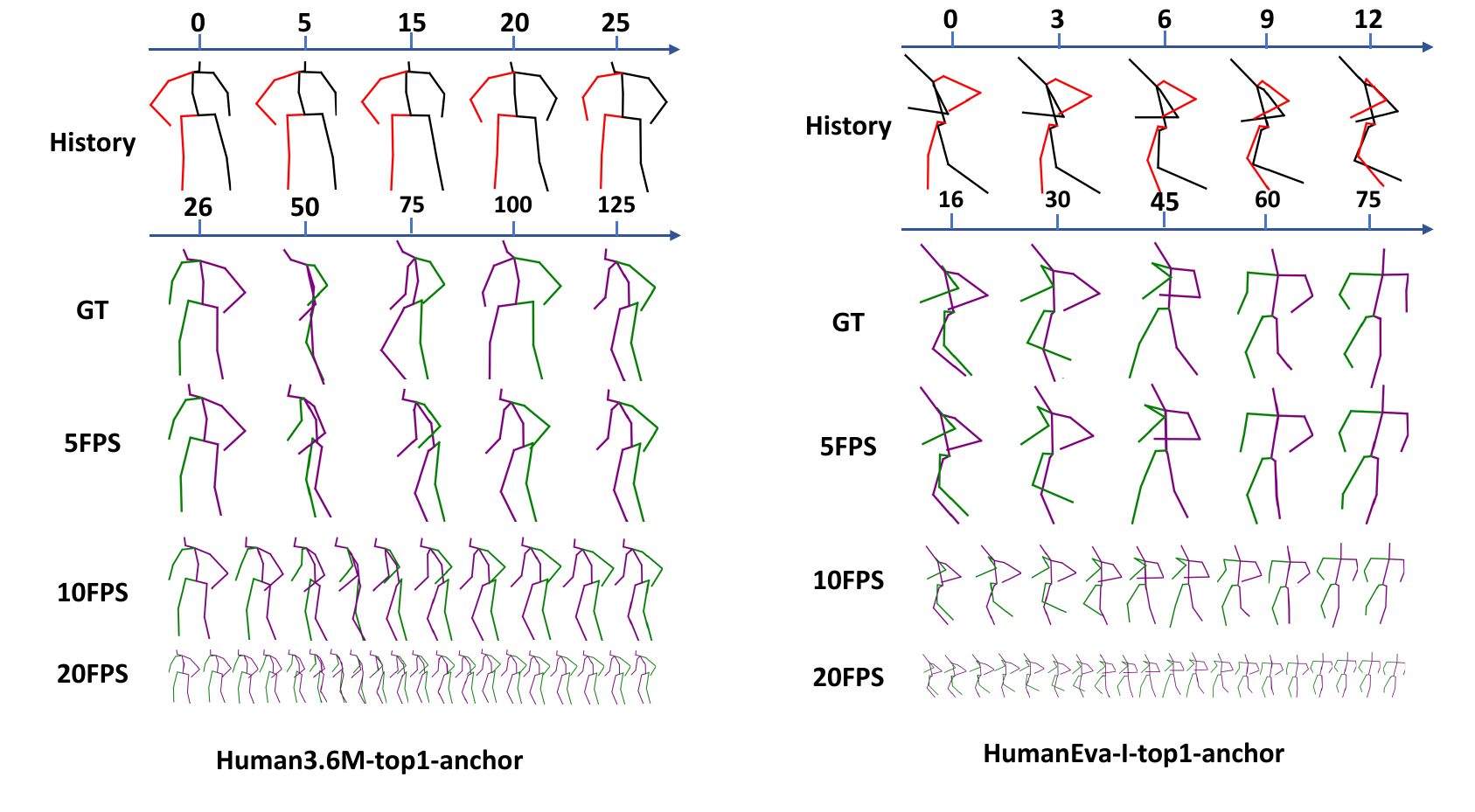}
    \caption{Visualization of the influence of spatio-temporal continuous network on different datasets. (The left of the Figure is sampled from the Human3.6M dataset. The right of the Figure is sampled from the HumanEva-I dataset). }
    \label{fig: ODE}
\end{figure*}

In this work, NeuralODE is mainly utilized to enhance the smoothness of predicted diverse human motion sequences, thereby facilitating the prediction accuracy of human motions. Therefore, in our experiments, we utilized the Mean Absolute Error (MAE) metric to report the influence of different ODE Solvers on the short-term and long-term predictions. Considering the diversity of numerical methods for solving ODEs, we tested both fixed-step and adaptive-step methods, and the results are presented in Table \ref{NeuralODE}.
In general, higher-order ODE Solvers tend to offer greater accuracy. However, it is important to note that higher orders often come with increased computational resource requirements. Nevertheless, when considering higher-order methods, adaptive-step methods still demonstrate superior performance compared to fixed-step methods.
Taking everything into account, Dopri5 consistently exhibited higher performance across all timestamps, making it the chosen ODE Solver for STCN.
\begin{figure*}[htbp]
    \centering
    \includegraphics[width=5.5in]{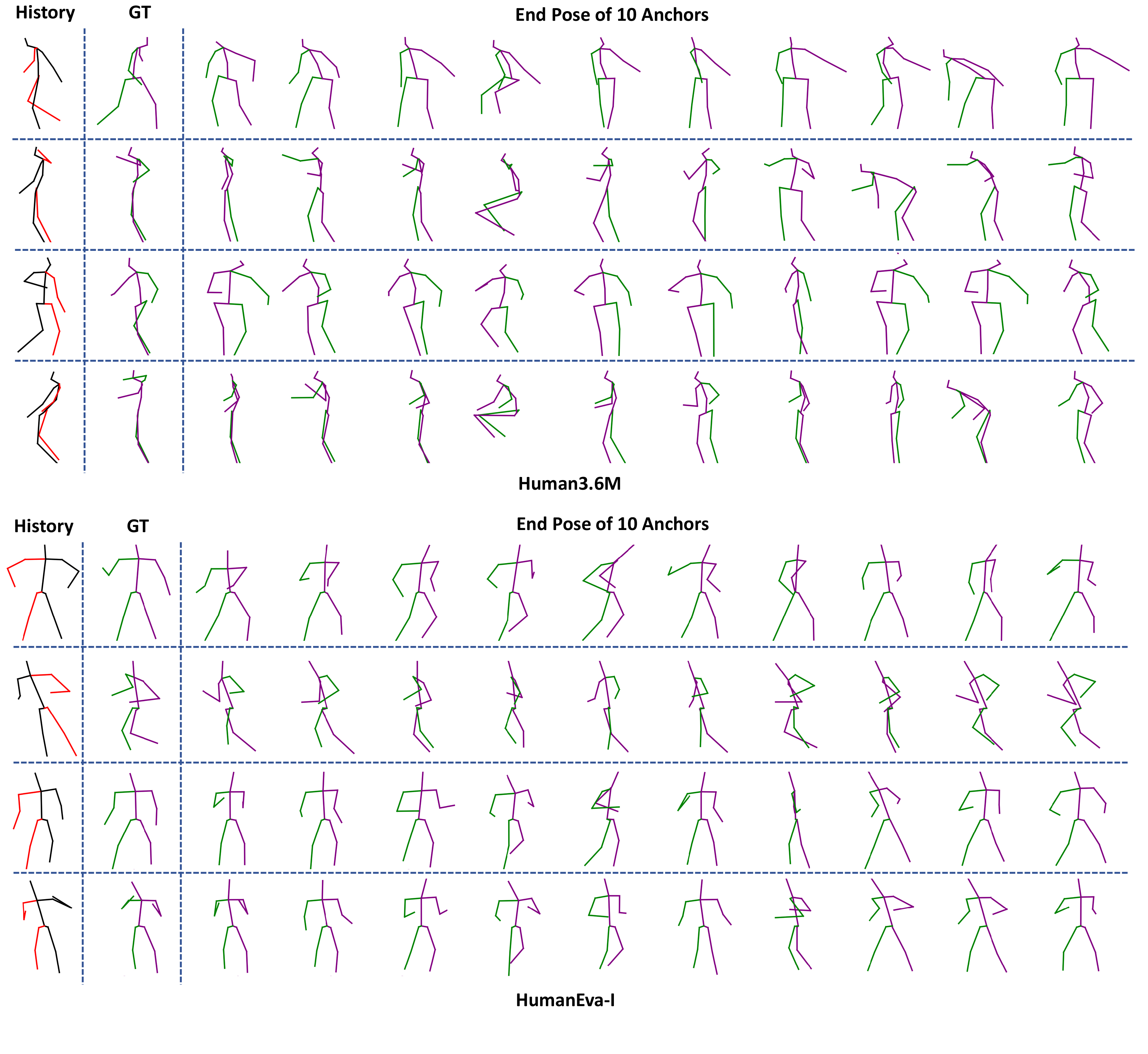}
    \caption{Visualization of end pose of 10 anchors for different human motion sequences on the Human3.6M (top) and HumanEva-I (bottom) datasets. }
    \label{fig: ResultSample}
\end{figure*}

\begin{figure*}[htbp]
    \centering
    \includegraphics[width=5.5in]{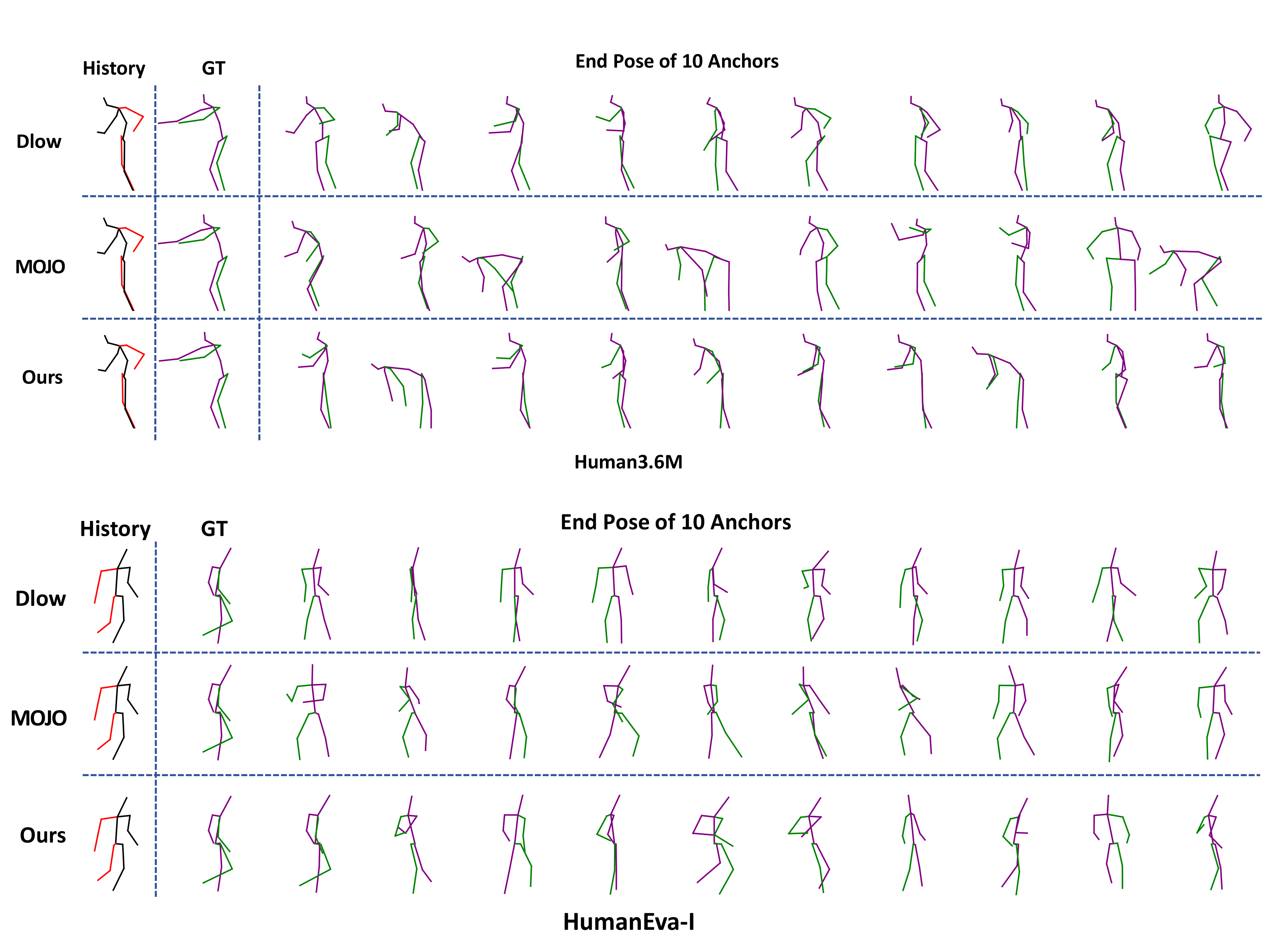}
    \caption{Comparison results of other popular methods and the proposed method on the Human3.6M and HumanEva-I datasets. }
    \label{fig: DiffMethod}
\end{figure*}

\subsection{Qualitative Results}
To qualitatively show the influence of different anchors on the performance, we visualize the number of anchors on the attention towards human body joints with the corresponding human body joints. The comparison results are illustrated in Fig.~\ref{fig:attention}. 
In Fig. 4 (left), we show the 16-joint skeleton used in the HumanEva‑I dataset, with the joints omitted (shown in parentheses) indicating the 17-joint skeleton used in the Human3.6M dataset, which is consistent with previous work \cite{yuan2020dlow}.
This figure denotes the skeleton-based human representation, which abstracts the human body as a set of joints and the limbs connecting these joints to form a skeleton. The skeletal are defined by this set of joints, and by constructing limbs between joints under specific constraints, the resulting skeleton adheres to a predefined kinematic tree structure (Kinematic Tree Structure).
The right side of this figure shows the influence of different anchors on the feature map. Fig.~\ref{fig:attention} shows that when setting the anchor to other values, the attention map remains relatively stable. However, significant changes occur in the attention maps when the anchor point is set to 20.
The changes in the feature map further demonstrate our choice.

We qualitatively evaluate the influence of the anchors by visualizing them in a low-dimensional space with t-SNE \cite{van2008visualizing}. As depicted in Fig.~\ref{fig:t-SNE}, different colors represent potential human motion patterns. The t-SNE depicted in the figure STCN is capable of automatically learning for clustering patterns. 
The human motion patterns exhibit some overlap, which is due to a few ambiguous actions that even humans find difficult to distinguish. These observations validate the advantages and the physical significance of the predefined anchors.

In addition, we also visualize the sampled sequences corresponding to different anchors. To exemplify, we employed motion data from the Human3.6M, as depicted in Fig.~\ref{fig:Sample}. To capture the fine-grained difference within the same anchor, multiple motion sequences are obtained by sampling from the GMM distribution. 
These results demonstrate the effectiveness of STCN in capturing intra-class differences in identical human motions, thereby enhancing the overall prediction accuracy.

Moreover, we show more qualitative results of STCN on HumanEva-I and Human3.6M datasets.
Specifically, Fig.~\ref{fig: ODE} illustrates the impact of the introduced spatio-temporal ODE-based network on the predicted human motion sequences. By sampling multiple motion sequences from the two human motion datasets at different Frames Per Second (FPS), we demonstrate the smoothness and coherence of the predicted human motion sequences.
These results provide further evidence of the effectiveness of our designed spatio-temporal ODE-based network in generating continuous and realistic human motion sequences. 

Fig.~\ref{fig: ResultSample} presents the predicted end poses of 10 samples for diverse human motion sequences. These predictions are generated based on different observed motion sequences and under various conditions in the HumanEva-I and Human3.6M datasets. The figure showcases the variability and diversity of the predicted human motion sequences.
In Fig.~\ref{fig: DiffMethod}, we provide a comparison of the diversity results between our method and other state-of-the-art techniques. The anchor is set to 10 for the comparison of other methods. It is evident that our method outperforms the baseline methods by generating stochastic and more realistic human motions. This comparison highlights the effectiveness and superiority of our approach in capturing the rich diversity of human motion data.

\section{Conclusion}
This paper proposes a novel method called STCN, to predict stochastic and continuous human motion sequences given the observed human motion sequences. STCN enhances the diversity and smoothness of predicted human motion sequences, which is achieved mainly by two stages.
In the first stage, STCN utilizes spatio-temporal
continuous network to learn continuous representations of dynamic human motions on the basis of the VQVAE network. 
Additionally, STCN innovatively introduces anchor set to capture the human motion patterns. 
In the second stage, STCN aims to capture the potential human motion patterns of the observed human motion sequence, and learn the Gaussian Mixture Model (GMM) distribution for each observed human motion sequence with the aid of the anchor sets. Simultaneously, the proposed method learns the probability scores and samples multiple sequences from each anchor to alleviate the intra-class differences in human motions. 
Comprehensive experimental results on the two widely used benchmark datasets demonstrate that STCN outperforms the state-of-the-art methods.

The proposed method STCN can enhance the smoothness and accuracy of the predicted stochastic human motion sequences, while alleviating the mode collapse.
STCN is easy to train compared with existing diffusion models and provides a novel method to enhance the performance of stochastic human motion prediction.
Given that our approach can be considered as the motion-conditioned generation, it can be extended to various conditional generation tasks, such as text-to-motion.
However, when the K-means algorithm is utilized to cluster the human motions, it has to predefine the number of anchors. Therefore, exploring unsupervised clustering algorithms could be a promising future direction for this method.

\bibliographystyle{unsrt}
\bibliography{cite}

\begin{thebibliography}{10}

\bibitem{yuan2023physdiff}
Ye~Yuan, Jiaming Song, Umar Iqbal, Arash Vahdat, and Jan Kautz.
\newblock Physdiff: Physics-guided human motion diffusion model.
\newblock In {\em Proceedings of the IEEE/CVF International Conference on
  Computer Vision}, pages 16010--16021, 2023.

\bibitem{tang2023collaborative}
Jin Tang, Jin Zhang, Rui Ding, Baoxuan Gu, and Jianqin Yin.
\newblock Collaborative multi-dynamic pattern modeling for human motion
  prediction.
\newblock {\em IEEE Transactions on Circuits and Systems for Video Technology},
  2023.

\bibitem{men2020quadruple}
Qianhui Men, Edmond~SL Ho, Hubert~PH Shum, and Howard Leung.
\newblock A quadruple diffusion convolutional recurrent network for human
  motion prediction.
\newblock {\em IEEE transactions on circuits and systems for video technology},
  31(9):3417--3432, 2020.

\bibitem{ding2022towards}
Pengxiang Ding and Jianqin Yin.
\newblock Towards more realistic human motion prediction with attention to
  motion coordination.
\newblock {\em IEEE Transactions on Circuits and Systems for Video Technology},
  32(9):5846--5858, 2022.

\bibitem{shen2022federated}
Qiang Shen et~al.
\newblock Federated multi-task attention for cross-individual human activity
  recognition.
\newblock In {\em IJCAI}, pages 3423--3429. IJCAI, 2022.

\bibitem{chen2023hdformer}
Hanyuan Chen, Jun-Yan He, Wangmeng Xiang, Wei Liu, Zhi-Qi Cheng, Hanbing Liu,
  Bin Luo, Yifeng Geng, and Xuansong Xie.
\newblock Hdformer: High-order directed transformer for 3d human pose
  estimation.
\newblock {\em arXiv preprint arXiv:2302.01825}, 2023.

\bibitem{liu2023multi}
Chenchen Liu and Yadong Mu.
\newblock Multi-granularity interaction for multi-person 3d motion prediction.
\newblock {\em IEEE Transactions on Circuits and Systems for Video Technology},
  2023.

\bibitem{yang2022motion}
Yang Yang, Guangjun Liu, and Xuehao Gao.
\newblock Motion guided attention learning for self-supervised 3d human action
  recognition.
\newblock {\em IEEE Transactions on Circuits and Systems for Video Technology},
  32(12):8623--8634, 2022.

\bibitem{hu2023personalized}
Junxing Hu, Hongwen Zhang, Yunlong Wang, Min Ren, and Zhenan Sun.
\newblock Personalized graph generation for monocular 3d human pose and shape
  estimation.
\newblock {\em IEEE Transactions on Circuits and Systems for Video Technology},
  2023.

\bibitem{zheng2022gimo}
Yang Zheng, Yanchao Yang, Kaichun Mo, Jiaman Li, Tao Yu, Yebin Liu, C~Karen
  Liu, and Leonidas~J Guibas.
\newblock Gimo: Gaze-informed human motion prediction in context.
\newblock In {\em European Conference on Computer Vision}, pages 676--694.
  Springer, 2022.

\bibitem{zhong2022spatio}
Chongyang Zhong, Lei Hu, Zihao Zhang, Yongjing Ye, and Shihong Xia.
\newblock Spatio-temporal gating-adjacency gcn for human motion prediction.
\newblock In {\em Proceedings of the IEEE/CVF Conference on Computer Vision and
  Pattern Recognition}, pages 6447--6456, 2022.

\bibitem{cai2021unified}
Yujun Cai et~al.
\newblock A unified 3d human motion synthesis model via conditional variational
  auto-encoder.
\newblock In {\em Proceedings of the IEEE/CVF International Conference on
  Computer Vision}, pages 11645--11655, 2021.

\bibitem{mao2021generating}
Wei Mao, Miaomiao Liu, and Mathieu Salzmann.
\newblock Generating smooth pose sequences for diverse human motion prediction.
\newblock In {\em Proceedings of the IEEE/CVF International Conference on
  Computer Vision}, pages 13309--13318, 2021.

\bibitem{liu2020trajectorycnn}
Xiaoli Liu, Jianqin Yin, Jin Liu, Pengxiang Ding, Jun Liu, and Huaping Liu.
\newblock Trajectorycnn: a new spatio-temporal feature learning network for
  human motion prediction.
\newblock {\em IEEE Transactions on Circuits and Systems for Video Technology},
  31(6):2133--2146, 2020.

\bibitem{chen2023spatiotemporal}
Haipeng Chen, Jiahui Hu, Wenyin Zhang, and Pengxiang Su.
\newblock Spatiotemporal consistency learning from momentum cues for human
  motion prediction.
\newblock {\em IEEE Transactions on Circuits and Systems for Video Technology},
  2023.

\bibitem{gui2018adversarial}
Gui et~al.
\newblock Adversarial geometry-aware human motion prediction.
\newblock In {\em Proceedings of the European Conference on Computer Vision
  (ECCV)}, pages 786--803, 2018.

\bibitem{odena2017conditional}
Odena et~al.
\newblock Conditional image synthesis with auxiliary classifier gans.
\newblock In {\em International conference on machine learning}, pages
  2642--2651. PMLR, 2017.

\bibitem{ma2021spatial}
Furong Ma, Guiyu Xia, and Qingshan Liu.
\newblock Spatial consistency constrained gan for human motion transfer.
\newblock {\em IEEE Transactions on Circuits and Systems for Video Technology},
  32(2):730--742, 2021.

\bibitem{barquero2023belfusion}
German Barquero, Sergio Escalera, and Cristina Palmero.
\newblock Belfusion: Latent diffusion for behavior-driven human motion
  prediction.
\newblock In {\em Proceedings of the IEEE/CVF International Conference on
  Computer Vision}, pages 2317--2327, 2023.

\bibitem{wei2023human}
Dong Wei, Huaijiang Sun, Bin Li, Jianfeng Lu, Weiqing Li, Xiaoning Sun, and
  Shengxiang Hu.
\newblock Human joint kinematics diffusion-refinement for stochastic motion
  prediction.
\newblock In {\em Proceedings of the AAAI Conference on Artificial
  Intelligence}, volume~37, pages 6110--6118, 2023.

\bibitem{yuan2020dlow}
Yuan et~al.
\newblock Dlow: Diversifying latent flows for diverse human motion prediction.
\newblock In {\em Computer Vision--ECCV 2020: 16th European Conference,
  Glasgow, UK, August 23--28, 2020, Proceedings, Part IX 16}, pages 346--364.
  Springer, 2020.

\bibitem{song2020score}
Yang Song, Jascha Sohl-Dickstein, Diederik~P Kingma, Abhishek Kumar, Stefano
  Ermon, and Ben Poole.
\newblock Score-based generative modeling through stochastic differential
  equations.
\newblock {\em arXiv preprint arXiv:2011.13456}, 2020.

\bibitem{van2017neural}
Van et~al.
\newblock Neural discrete representation learning.
\newblock {\em Advances in neural information processing systems}, 30, 2017.

\bibitem{barsoum2018hp}
Barsoum et~al.
\newblock Hp-gan: Probabilistic 3d human motion prediction via gan.
\newblock In {\em Proceedings of the IEEE conference on computer vision and
  pattern recognition workshops}, pages 1418--1427, 2018.

\bibitem{ho2020denoising}
Ho et~al.
\newblock Denoising diffusion probabilistic models.
\newblock {\em Advances in Neural Information Processing Systems},
  33:6840--6851, 2020.

\bibitem{zhang2022motiondiffuse}
Zhang et~al.
\newblock Motiondiffuse: Text-driven human motion generation with diffusion
  model.
\newblock {\em arXiv preprint arXiv:2208.15001}, 2022.

\bibitem{tevet2022human}
Guy Tevet, Sigal Raab, Brian Gordon, Yonatan Shafir, Daniel Cohen-Or, and
  Amit~H Bermano.
\newblock Human motion diffusion model.
\newblock {\em arXiv preprint arXiv:2209.14916}, 2022.

\bibitem{jiang2023motiongpt}
Biao Jiang, Xin Chen, Wen Liu, Jingyi Yu, Gang Yu, and Tao Chen.
\newblock Motiongpt: Human motion as a foreign language.
\newblock {\em arXiv preprint arXiv:2306.14795}, 2023.

\bibitem{ruiz2023neural}
Domenec Ruiz-Balet and Enrique Zuazua.
\newblock Neural ode control for classification, approximation, and transport.
\newblock {\em SIAM Review}, 65(3):735--773, 2023.

\bibitem{fang2021spatial}
Zheng Fang, Qingqing Long, Guojie Song, and Kunqing Xie.
\newblock Spatial-temporal graph ode networks for traffic flow forecasting.
\newblock In {\em Proceedings of the 27th ACM SIGKDD conference on knowledge
  discovery \& data mining}, pages 364--373, 2021.

\bibitem{xing2023hdg}
Yucheng Xing and Xin Wang.
\newblock Hdg-ode: A hierarchical continuous-time model for human pose
  forecasting.
\newblock In {\em Proceedings of the IEEE/CVF International Conference on
  Computer Vision}, pages 14700--14712, 2023.

\bibitem{chen2024neural}
Yang Chen, Hong Liu, Pinhao Song, and Wenhao Li.
\newblock Neural ordinary differential equation for irregular human motion
  prediction.
\newblock {\em Pattern Recognition Letters}, 178:76--83, 2024.

\bibitem{chai2019multipath}
Chai et~al.
\newblock Multipath: Multiple probabilistic anchor trajectory hypotheses for
  behavior prediction.
\newblock {\em arXiv preprint arXiv:1910.05449}, 2019.

\bibitem{erhan2014scalable}
Dumitru Erhan, Christian Szegedy, Alexander Toshev, and Dragomir Anguelov.
\newblock Scalable object detection using deep neural networks.
\newblock In {\em Proceedings of the IEEE conference on computer vision and
  pattern recognition}, pages 2147--2154, 2014.

\bibitem{ionescu2013human3}
Ionescu et~al.
\newblock Human3. 6m: Large scale datasets and predictive methods for 3d human
  sensing in natural environments.
\newblock {\em IEEE transactions on pattern analysis and machine intelligence},
  36(7):1325--1339, 2013.

\bibitem{sigal2010humaneva}
Sigal et~al.
\newblock Humaneva: Synchronized video and motion capture dataset and baseline
  algorithm for evaluation of articulated human motion.
\newblock {\em International journal of computer vision}, 87(1-2):4, 2010.

\bibitem{mao2019learning}
Wei Mao, Miaomiao Liu, Mathieu Salzmann, and Hongdong Li.
\newblock Learning trajectory dependencies for human motion prediction.
\newblock In {\em Proceedings of the IEEE/CVF International Conference on
  Computer Vision}, pages 9489--9497, 2019.

\bibitem{dang2021msr}
Lingwei Dang, Yongwei Nie, Chengjiang Long, Qing Zhang, and Guiqing Li.
\newblock Msr-gcn: Multi-scale residual graph convolution networks for human
  motion prediction.
\newblock In {\em Proceedings of the IEEE/CVF International Conference on
  Computer Vision}, pages 11467--11476, 2021.

\bibitem{li2018structure}
Xiu Li, Hongdong Li, Hanbyul Joo, Yebin Liu, and Yaser Sheikh.
\newblock Structure from recurrent motion: From rigidity to recurrency.
\newblock In {\em Proceedings of the IEEE conference on computer vision and
  pattern recognition}, pages 3032--3040, 2018.

\bibitem{walker2017pose}
Jacob Walker, Kenneth Marino, Abhinav Gupta, and Martial Hebert.
\newblock The pose knows: Video forecasting by generating pose futures.
\newblock In {\em Proceedings of the IEEE international conference on computer
  vision}, pages 3332--3341, 2017.

\bibitem{yan2018mt}
Xinchen Yan, Akash Rastogi, Ruben Villegas, Kalyan Sunkavalli, Eli Shechtman,
  Sunil Hadap, Ersin Yumer, and Honglak Lee.
\newblock Mt-vae: Learning motion transformations to generate multimodal human
  dynamics.
\newblock In {\em Proceedings of the European conference on computer vision
  (ECCV)}, pages 265--281, 2018.

\bibitem{dilokthanakul2016deep}
Nat Dilokthanakul, Pedro~AM Mediano, Marta Garnelo, Matthew~CH Lee, Hugh
  Salimbeni, Kai Arulkumaran, and Murray Shanahan.
\newblock Deep unsupervised clustering with gaussian mixture variational
  autoencoders.
\newblock {\em arXiv preprint arXiv:1611.02648}, 2016.

\bibitem{yuan2019diverse}
Ye~Yuan et~al.
\newblock Diverse trajectory forecasting with determinantal point processes.
\newblock {\em arXiv preprint arXiv:1907.04967}, 2019.

\bibitem{zhang2021we}
Yan Zhang et~al.
\newblock We are more than our joints: Predicting how 3d bodies move.
\newblock In {\em Proceedings of the IEEE/CVF Conference on Computer Vision and
  Pattern Recognition}, pages 3372--3382, 2021.

\bibitem{dang2022diverse}
Lingwei Dang, Yongwei Nie, Chengjiang Long, Qing Zhang, and Guiqing Li.
\newblock Diverse human motion prediction via gumbel-softmax sampling from an
  auxiliary space.
\newblock In {\em Proceedings of the 30th ACM International Conference on
  Multimedia}, pages 5162--5171, 2022.

\bibitem{van2008visualizing}
Laurens Van~der Maaten and Geoffrey Hinton.
\newblock Visualizing data using t-sne.
\newblock {\em Journal of machine learning research}, 9(11), 2008.

\end{thebibliography}

\begin{IEEEbiography}[{\includegraphics[width=1in,height=1.25in,clip,keepaspectratio]{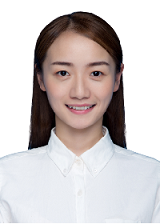}}]{Hua Yu}
is currently pursuing Ph.D. degree from Dalian University Of Technology in China and Nanyang Technological University in Singapore. Her research interests include human-robot interactions, human motion understanding, and artificial intelligence. Currently, she serves as the reviewer of the IEEE Transactions on Circuits and Systems for Video Technology, ACM Multimedia, and IEEE Transactions on Cybernetics.
\end{IEEEbiography}

\begin{IEEEbiography}[{\includegraphics[width=1in,height=1.25in,clip,keepaspectratio]{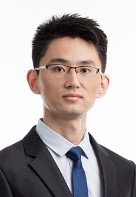}}]{Yaqing Hou} received the Ph.D. degree in artificial intelligence from the Interdisciplinary Graduate School, Nanyang Technological University, Singapore, in 2017. He was a Post-Doctoral Research Fellow with the Data Science and Artificial Intelligence Research Centre, Nanyang Technological University. 
He is currently an Associate Professor with the College of Computer Science and Technology, Dalian University of Technology, Dalian, China. He is Associate Editor of the Memetic Computing. His current research interests include computational and artificial intelligence, memetic computing, multi-agent reinforcement learning, transfer learning and optimization.
\end{IEEEbiography}

\begin{IEEEbiography}[{\includegraphics[width=1in,height=1.25in,clip,keepaspectratio]{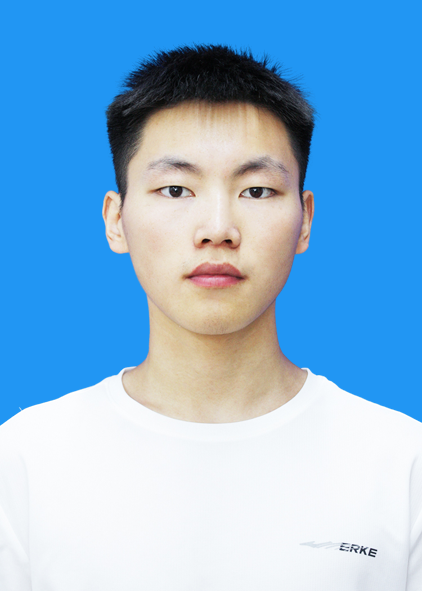}}]{Xu Gui}
received the B.S from Sichuan Normal University, China, in 2022. He is currently a master student in the School of Software Engineering, Dalian University (DLU), China. His current research interest consists of human motion prediction and human motion understanding.
\end{IEEEbiography}

\begin{IEEEbiography}[{\includegraphics[width=1in,height=1.25in,clip,keepaspectratio]{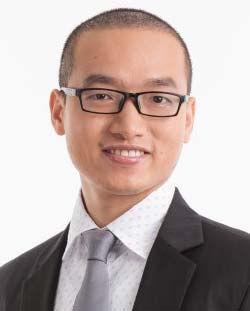}}]{Shanshan Feng}
received the B.S. degree from the University of Science and Technology of China in 2011, and the Ph.D. degree from Nanyang Technological University in 2017. He is currently the chief scientist at Wecar Technology Company, Ltd., Shenzhen, China. Before that, he worked at the School of Computer Science and Technology, Harbin Institute of Technology (Shenzhen), and Inception Institute of Artificial Intelligence (IIAI), Abu Dhabi, United Arab Emirates. His current research interests include big data analytics, spatiotemporal data mining, artificial intelligence, and graph learning.
\end{IEEEbiography}

\begin{IEEEbiography}[{\includegraphics[width=1in,height=1.25in,clip,keepaspectratio]{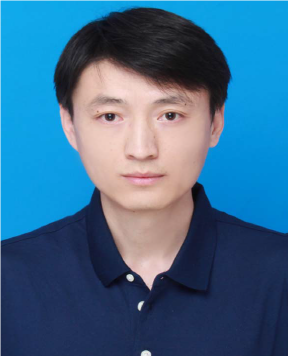}}]{Dongsheng Zhou} (Member, IEEE) was born in 1978. He received the Ph.D. degree from the Dalian University of Technology. He is currently a Distinguish Professor of Liaoning Province. His research interests include CG, intelligence computing, and human–robot interaction. He is a member of ACM, CGS, and CCF.
\end{IEEEbiography}

\begin{IEEEbiography}[{\includegraphics[width=1in,height=1.25in,clip,keepaspectratio]{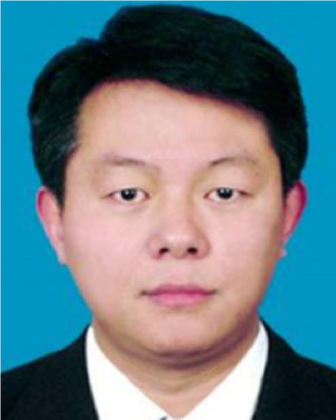}}]{Qiang Zhang} was born in Xi’an, China, in 1971. He received the M.Eng. degree in economic engineering and Ph.D. degree in circuits and systems from Xidian University, Xi’an, in 1999 and 2002, respectively. In 2003, he was a Lecturer with the Center of Advanced Design Technology, Dalian University, Dalian, China, where he was a Professor in 2005. His research interest includes bio-inspired computing and its applications. He has authored more than 70 articles in the above ﬁelds. Thus far, he has served on the editorial board of seven international journals and has edited special issues in journals, such as Neurocomputing and the International Journal of Computer Applications in Technology.
\end{IEEEbiography}

\end{document}